# Attention Based Encoder Decoder Model for Video Captioning in Nepali (2023)

Kabita Parajuli, *Department of Electronics and Computer Engineering, Pulchowk Campus, Tribhuvan University,*
Shashidhar R. Joshi, *Department of Electronics and Computer Engineering, Pulchowk Campus, Tribhuvan University*

*Abstract*—Video captioning in Nepali, a language written in the Devanagari script, presents a unique challenge due to the lack of existing academic work in this domain. This work develops a novel encoder-decoder paradigm for Nepali video captioning to tackle this difficulty. LSTM and GRU sequence-to-sequence models are used in the model to produce related textual descriptions based on features retrieved from video frames using CNNs. Using Google Translate and manual post-editing, a Nepali video captioning dataset is generated from the Microsoft Research Video Description Corpus (MSVD) dataset created using Google Translate, and manual post-editing work. The efficiency of the model for Devanagari-scripted video captioning is demonstrated by BLEU, METOR, and ROUGE measures, which are used to assess its performance.

## I. INTRODUCTION

The proliferation of multimedia data, particularly videos, has yielded a number of advantages but also presented challenges in terms of organising and obtaining access to the vast amount of available visual data. The widespread use of the internet of video material has led to the importance of video captioning as a topic of research.Effectively organising, indexing, and retrieving videos is essential to handling and understanding this massive amount of visual information. The increasing popularity of websites that exchange films has led to a desire for accurate and effective ways for comprehending videos.

Video captioning is the technique of automatically generating a description of a video in natural language. It is a challenging task because videos are dynamic and complicated. Unlike photos, videos have a temporal component, which implies that they change over time. This makes it difficult for models to extract the temporal and spatial information needed to provide correct and informative captions. Statistical language models are sometimes combined with handcrafted features like as motion vectors or low-level visual descriptors in conventional approaches. Transformer models or recurrent neural networks (RNNs) are employed in deep learning-based techniques for language modelling, and convolutional neural networks (CNNs) are used for visual analysis.

Deep learning-based approaches have recently set the standard for video captioning. These techniques use convolutional neural networks (CNNs) to extract visual data from the video, after which a caption is produced utilising recurrent neural networks (RNNs) or transformer models based on these data. Because CNNs are very good at learning complex spatial patterns, they are particularly useful for extracting visual data from movies. RNNs are perfect for caption creation because they can record the temporal relationships between words. Transformer models, a more modern kind of neural networks, have shown promising results in video captioning. Their attention system allows them to swiftly understand long-range dependencies, which makes them effective at occupations like video captioning.

While LSTM and GRU are two popular recurrent neural network (RNN) architectures used for video captioning, each has pros and cons of its own. LSTM is an excellent choice for tasks requiring long-term memory, such as video captioning, because of its exceptional ability to capture long-term dependencies in data. However, higher complexity has a higher processing overhead; conversely, a simpler structure speeds up inference and training and boosts processing efficiency. But given its simplicity, it might not be as good at simulating long-term dependence. The demands of the current task determine which of LSTM and GRU to use. Even if it means sacrificing speed, LSTM is the best choice for applications where precision is crucial. Even at the expense of some precision, GRU performs better on jobs that call for real-time execution. When speed and accuracy are equally important, a hybrid strategy that combines LSTM and GRU might be the best choice.

## II. LITERATURE REVIEW

Nepali is one of the notable exceptions to the rule that most video captioning has been completed for a limited number of widely spoken languages. Inspired by advances in image captioning, researchers have experimented with a range of deep learning architectures and frameworks to generate captions from films or image frame sequences. The basis of previous approaches was the prediction of subject-verb-object (SVO) relationships and matching them to pre-established templates. Researchers proposed methods that linked motion verbs to vehicle movements to allow descriptions of vehicle actions. Semantic tags were added to the description creation process to better enhance it. Through object observation in naturalistic scenarios, they established 64 verbs of motion to characterise the motion of a car. By developing semantic tags for the behaviours displayed in the film, the process of generating descriptions was extended even further. Recently, the focus of researchers has changed.

after examining the remarkable achievements of deep learning (DL) in the domains of both CV and NLP, away from utilising preset templates for generating descriptions in favour of deep recurrent networks. Today, the majority of video captioning techniques use an encoder-decoder framework that makes use of different CNN (Convolutional Neural Network) configurations and variations of recurrent networks.

L. Yan et al. presented a method for merging global and local representations to create descriptive and contextually relevant video subtitles. [1] Global representation is captured in the first stream by extracting features from the entire video sequence, which yields high-level knowledge. The second stream focuses on local representations by detecting regions inside video frames through feature extraction. These global and local data are subsequently provided to a caption generator that makes use of transformer-based designs or recurrent neural networks. The outcomes of their trials demonstrate that, in terms of relevancy and caption quality, their method outperforms existing methods.

Sandeep Samleti et al. [6] developed a video captioning system that leverages an LSTM for sequence synthesis, a CNN for frame-level feature extraction, and a mean-pooled vector of all recovered features to represent the entire movie. Regretfully, this method is unable to capture the temporal correlations between frames due to mean pooling. Kevin Lin et al. [5] built on Samleti's work by introducing a two-layer LSTM encoder-decoder architecture for video captioning. Every frame is used as input to construct a fixed-size feature vector comprising visual features at each time step.

An intriguing study in this area titled "Attention-based video captioning framework for Hindi" was published later this year. Alok Singh et al.'s [2] attempt to address the captioning problem for Hindi videos. In a language-rich nation such as India, providing a way for native speakers to understand the visual entities is essential. To enable the system to choose the right moment to focus on both visual context vector and semantic information, we use a hybrid attention mechanism in this study that combines a soft temporal attention mechanism with a semantic attention mechanism. The visual context vector of the input video is extracted using a 3D convolutional neural network (3D CNN), and then the encoded context vector is decoded using a Long Short-Term Memory (LSTM) recurrent network with an attention module. An internal dataset designed for Hindi video captioning was used to translate and post-edit the (MSR-VTT) dataset. This model outperforms other baseline models with a 0.369 CIDEr score and a 0.393 METEOR score. For an input video, this model was unable to produce subtitles in many languages.

The object relation graph and multimodal feature fusion serve as the foundation for Zhiwen Yan et al.'s groundbreaking video captioning system (ORMF) [3]. Graph convolution network (GCN) is introduced to encode the object relation using ORMF, which builds an object relation features graph based on the spatiotemporal correlations and similarities among the objects in the film. To determine how various modal features relate to one another, ORMF builds a multi-modal features fusion network. Many modalities' features are combined in the multimodal feature fusion network. By calculating the length loss of the caption, the suggested model generates a richer caption. The experimental findings on two publicly

The benefit of this approach is demonstrated by the available datasets (Microsoft research-video to text [MSR-VTT]).

The essay "SBAT: Video Captioning with Sparse Boundary-Aware Transformer" [4] by Tao Jin et al. focuses on using the transformer structure for video captioning. For unimodal language-generating applications, such machine translation, the vanilla transformer is advised. Since the video data contains a lot of overlap between different time steps, video captioning presents a multimodal learning difficulty. Taking these issues into account, we suggest a novel technique known as the sparse boundary-aware transformer (SBAT) to eliminate redundant information in video representation. After choosing a variety of characteristics from various contexts, SBAT applies a boundary-aware pooling technique to multi-head attention scores. Additionally, SBAT has a local correlation strategy to make up for the local information loss caused by sparse operation. Using the MSVD dataset, SBAT outperforms TVT and MARN models overall and outperforms GRU-EVE, SCN, and POS-CG in every category. As attention weights are assigned, the vanilla transformer's capacity to recognise the boundaries of various scenarios decreases.

These days, the majority of video captioning methods use on encoder-decoder frameworks that combine convolutional neural networks and several recurrent network versions. A video captioning system was given by [17] that employs CNN to extract frame-level features. A mean-pooled vector of all the gathered features is then utilised to represent the entire video. After that, an LSTM receives this vector to produce sequences. The disadvantage of this is that the mean pooling method is unable to capture the temporal link between the frames. Building on previous work, [18] proposed a second encoder-decoder method using two layers of LSTMs for video captioning. This framework uses each frame as an input in each time step to encode the video's visual elements into a fixed size feature vector.

LSTM (Long Short-Term Memory) and GRU (Gated Recurrent Unit) models are very useful tools for video captioning tasks because of their capacity to handle sequential data and long-term dependencies. When it comes to gathering and preserving significant information from movies, these models perform remarkably well, especially when contextual dependencies span multiple frames or time steps. Their intricate internal mechanisms let them to effectively update and preserve relevant data over long periods of time, enabling them to encode and capture the temporal dynamism inherent in movies. Additionally, because videos arrive in a range of lengths, the LSTM and GRU models can easily handle sequences with varying lengths. This is a crucial aspect for video captioning.

Previous research has focused on key languages such as Hindi, Chinese, English, and German. This gave me the idea to use the MSVD dataset, which is freely accessible and has been translated into Nepali. I then post-edited each reference caption in an effort to address the issue of video captioning in the NEPALI language. A machine translation system can translate an English caption into Nepali, or it can train a model directly on a Nepali reference caption to generate output in Nepali. The generated captions' quality will deteriorate if the

They are generated in a variety of languages using the MT approach. Google Translate is my preferred method of manual translation.

### A. Long Short-Term Memory Architecture

Long short-term memory is an artificial recurrent neural network architecture that blends feedforward and feedback neural networks. Because of their special architecture, which enables them to maintain the relationship between recent past knowledge and current tasks even as the gap increases, LSTMs solve long-term dependency. The LSTM structure is a type of memory system that can discriminate between information that should be disseminated and information that should be retained. Information travels between cells in this structure. The cell state can carry information regarding sequential data processing, which includes text, speech, video, and other media.

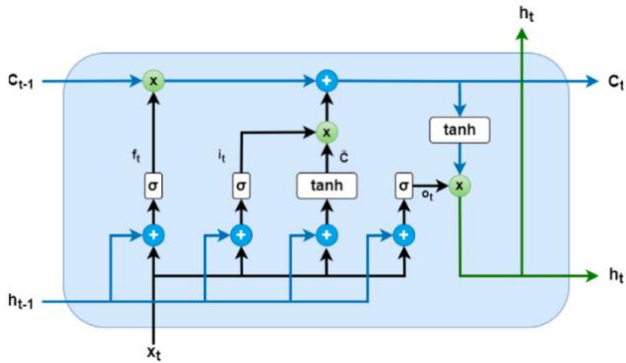

**Fig 1**: Basic Architecture of LSTM

A cell state is made up of the input gate ($x_t$), output gate ($h_t$), and forget gate ($f_t$). An LSTM cell uses an input gate to determine how important fresh information is at a given timestamp. A forget gate is the essential component responsible for removing (forgetting) the data from the previous timestamp. Furthermore, selecting the most significant data from the running LSTM cell and sending it out as the output is the aim of the output gate.

$$i_t = \sigma(x_i U^i + h_{t-1} W^i) \quad (1)$$
$$f_t = \sigma(x_t U^f + h_{t-1} W^f) \quad (2)$$
$$o_t = \sigma(x_t U^o + h_{t-1} W^o) \quad (3)$$
$$C'_t = \tanh(x_t U^g + h_{t-1} W^g). \quad (4)$$
$$C_t = \tanh(x_t U^g + h_{t-1} W^g). \quad (5)$$
$$h_t = \tanh(C_t) * o_t \quad (6)$$

The mathematical formulas for the corresponding gates in the LSTM architecture are shown in equations 1 through 6. The adoption of LSTM in this work is justified by its capacity to selectively hold onto patterns for long epochs. Furthermore, the LSTM design facilitates the efficient classification, processing, and accurate forecasting of large amounts of time-series data.

### B. Gated Recurrent Unit

A gated recurrent unit (GRU) is a simple recurrent neural network with a gating mechanism added. Similar to LSTM, In GRU, gates are used to control the flow of information. Compared to LSTM, it can train faster and has a simpler, parameter-light design. An update gate ($z_t$), reset gate ($r_t$), current memory content ($h_t$), and final memory at the current time step ($h_t$) comprise the basic architecture of a single GRU unit, as depicted in Figure. The following are the GRU mathematical formulas:

$$z_t = W_z x x_t + W_z h h_t + b_z \quad (7)$$
$$r_t = (W_r x x_t + W_r h h_{t-1} + b_r) \quad (8)$$
$$h'_t = \tanh(W_{\sigma x} x_t + W_{\sigma h} h_{t-1} + b_\sigma) \quad (9)$$
$$h_t = z_t h_{t-1} + (1 - z_t) h'_t \quad (10)$$

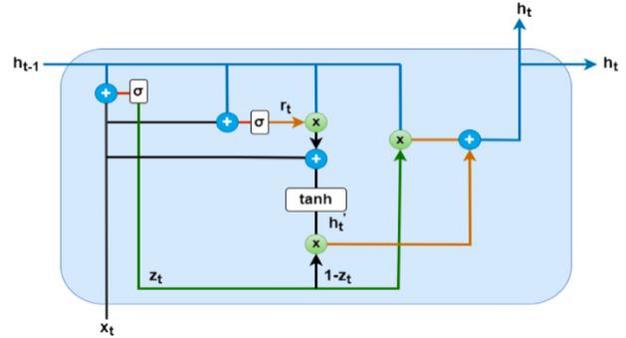

**Fig 2**: Basic Architecture of GRU

### C. Encoder

The encoder's goal is to decipher and analyse the input sequence. Whether the input is a sequence of video frames or a sentence in natural language, the encoder steps through the input, extracting and evaluating relevant information at each stage. In text-based tasks, every word in the sentence or every token in the input sequence is analysed one after the other. The encoder's LSTM or GRU units modify their internal hidden states in response to inputs at each time step. These hidden states take important information out of the input sequence. The encoder's final hidden state, known as the context vector or thinking vector, comprises a compressed representation of the whole input sequence. It is noteworthy that in numerous applications, the encoder's output is removed and only the internal states are retained.

The encoder cell architecture might vary depending on the type of recurrent neural network (RNN) being used. One of the primary differences between GRU (Gated Recurrent Unit) and LSTM (Long Short-Term Memory) is the internal state setup. An LSTM cell has two internal states: the cell state, which is in charge of long-term data management, and the hidden state, which holds information carried over from one time step to the next. GRU cells, on the other hand, contain a single hidden state, making their construction simpler.

Using an attention mechanism also shows a significant departure from traditional techniques. Instead of focusing only on the last step, this method takes into account the outputs from all 80 encoder states. This shift in attention allows the model to flexibly attend to different regions of the input sequence, which enhances its capacity to extract and apply relevant information for the task at hand. Sequence-to-sequence models' effectiveness is strongly impacted by this architecture and design, particularly in the domains of natural language processing and other disciplines that depend on comprehension and context.

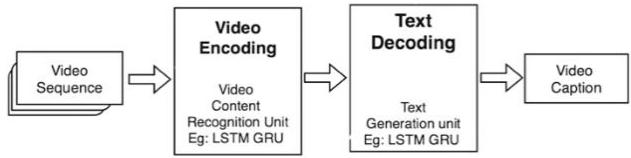

**Fig** 3: Basic Architecture of Encoder Decoder

*D. Decoder*

The encoder is followed by the decoder, which generates the output sequence. Its initial LSTM or GRU cell is initialised using the encoder's context vector. Creating the output sequence piece by piece is the primary duty of the decoder. For jobs involving text synthesis, such language translation or text summarization, the decoder generates words one at a time. When used to caption videos, it generates intelligent subtitles for every frame. The decoder's LSTM or GRU cell's initial hidden state, which acts as the basis for generating the output, is established using the context vector. At each time step, the decoder generates an output token (a word, for example) based on the previous output and the hidden state it is currently in.

The decoder cells take in the vector representation of the reference captions in order to learn the mapping function. The <start> (sentence starting) token is used in the decoder's first cell to forecast the first vector. Similarly, the output of the first decoder cell is used by the second cell to forecast the second vector, and so on. The process continues until the <end> (end of sentence) token is encountered.

Since both the encoder and the decoder employ LSTM or GRU cells, the encoder-decoder may effectively capture and use sequential dependencies. By storing hidden states containing pertinent details about the sequence that has been processed thus far, these recurrent units allow the model to consider context while generating predictions. When it comes to video captioning, this architecture allows the model to take in a sequence of video frames and provide captions that make sense within the context. The context vector that the encoder creates is crucial to ensuring that the captions correspond with the content of the video frames. Let us conclude that the encoder-decoder architecture with LSTM/GRU cells provides a strong foundation for sequence-to-sequence operations. It uses the decoder to generate matching output sequences and the encoder to understand input sequences, making it an adaptable tool for many uses, such as language translation and video captioning.

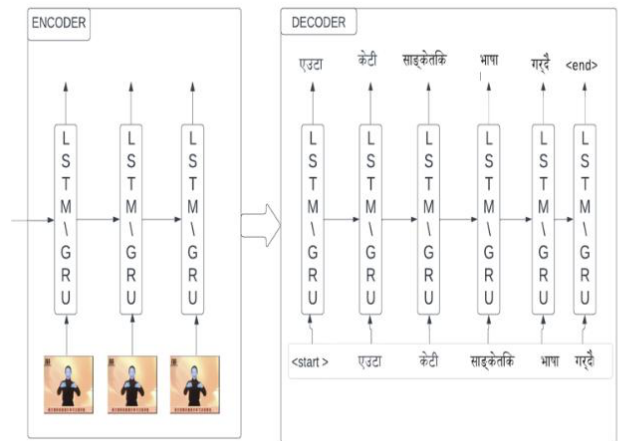

**Fig** 4: Working of Encoder Decoder model during Training

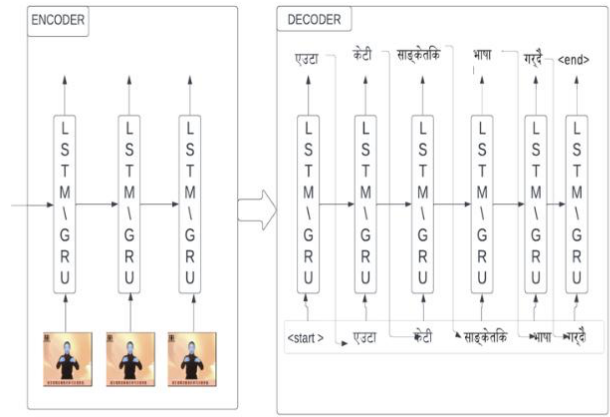

**Fig** 5: Working of Encoder Decoder model during Prediction

*E. Attention Mechanism*

One limitation of the encoder-decoder-based seq2seq design is that the entire input sequence is encoded into a single fixed-length character vector. Working with extremely long input sequences—especially ones longer than those in the training dataset—presents this problem. The attention model, which generates a context vector specific to each decoder time step, was introduced to get around this restriction. This approach translates the entire input sequence onto a single context vector, which sets it apart from the standard encoder-decoder paradigm.

The input and output components both influence the attention mechanism. The decoder can employ the most relevant information from the input sequence in a flexible way thanks to attention. In order to do this, all of the encoder outputs are combined into a weighted combination, with the highest weights going to the vectors that are most relevant. This attention mechanism is used to compare the two models (GRU, LSTM) that were previously mentioned and evaluate which performs better. This problem was overcome with the inclusion of the attention model, which produced a context vector that was distinct from the standard at every decoder time step.

III. METHODOLOGY

Video captioning using an encoder-decoder paradigm is a powerful, flexible technology with a wide range of applications. It is an illustration of how artificial intelligence (AI) may bridge gaps between different media forms (textual and visual) and create more inclusive and instructive digital experiences. As the discipline develops further, it offers exciting prospects to improve our understanding of and interaction with video information in an increasingly digitised world. Video captioning using an encoder-decoder architecture is an exciting and multidisciplinary field of research that combines computer vision and natural language processing. With its potential to make movies more accessible, findable, and useful, this tactic is gaining a lot of attention and has a wide range of real-world applications. By enabling machines to understand and describe video information, it opens up new possibilities.

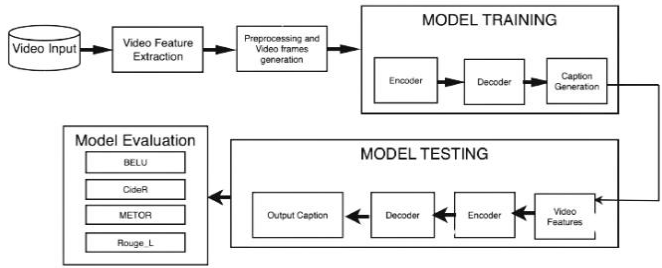

**Fig** 6: Proposed model block diagram

*A. Dataset*

Prior to work, there is no dataset available for Nepali video captioning. Therefore, the Microsoft Research Video Description (MSVD) Dataset includes over 2,000 short videos, each lasting a few seconds and covering a broad range of subjects and activities. Each video also includes fifteen to twenty subtitles. An enormous collection of YouTube video clips and the accompanying annotations may be found in the MSVD dataset. Across 1,970 videos lasting between 10 and 20 seconds, it contains over 80,789 video description sentences. Using Google Translate, a special Nepali video captioning dataset is generated from the MSVD dataset.

*B. Preprocessing*

There are a few challenges with using Google Translate to translate an English caption into Nepali, such as long and ambiguous captions being translated wrongly. After translation, we made the translated captions cleaner by hand editing them. Every caption is tokenized using a Nepali tokenizer..

*C. Feature Extraction*

In order to extract features for video captioning using the MSVD (Microsoft Research Video Description) dataset, it is necessary to collect both temporal and visual data from the video frames.

insightful captions. The VGG16 model is used to extract frames from videos. The video path and the required number of extracted frames are two arguments that the frames function needs in order to extract frames from a given video file. To ensure that representative frames are recorded, the function reads frames and distributes them evenly across the movie. a comprehensive pipeline for prepping videos that gathers frames from a movie dataset, stores them as NumPy arrays, and stores the captions associated with each frame..

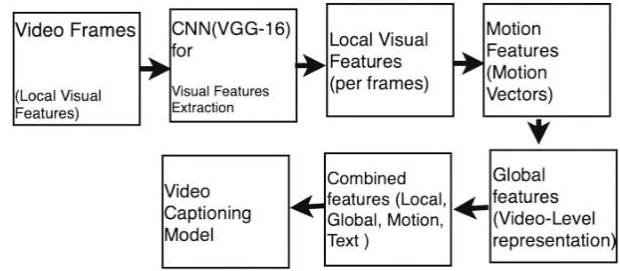

**Fig** 7: Feature Extraction overall illustration

During the pre-processing phase, video frames are retrieved and any required modifications are applied. Use the extract frame function to extract frames from a specified video file. It needs some parameters, like the required amount of extracted frames and the video route. To ensure that representative frames are recorded, the function reads frames and distributes them evenly across the movie. a comprehensive pipeline for prepping videos that gathers frames from a movie dataset, stores them as NumPy arrays, and stores the captions associated with each frame. For every selected frame, the pre-trained model extracts features. When a large dataset with millions of images is used to pre-train a model such as VGG16, the result is a vector of features, often 4096 in size, that captures the salient visual characteristics of each frame as the model processes it. After these feature vectors for every film are stacked, a structured NumPy array with size (28, 4096) is generated.

```
[[1.4559064  0.66230696 0.08930378 ... 0.25029743 1.162832   1.2205805 ]
 [0.62838507 0.6127548  0.15510947 ... 0.45860204 1.2756816  1.3812673 ]
 [0.7599578  0.58748186 0.26961824 ... 0.44239733 1.3105797  1.1350001 ]
 ...
 [0.8577998  0.24253726 0.16858245 ... 0.4667028  0.24543582 0.6936421 ]
 [0.73376036 0.55585486 0.2657315  ... 0.04579229 0.3486544  0.7601513 ]
 [0.74133354 0.67648816 0.4922339  ... 0.27372175 1.1050256  0.8400312 ]]
```

**Fig** 8: Structured NumPy Array of dimensions (28,4096)

These arrays offer a format that captures the essential visual elements of the video content for machine-learning algorithms. By extracting these elements, the dataset becomes a feature-rich representation that may be used as an input to generate captions that explain the video's content. Global features are extracted from local data and represent an overview or combination of the local features that encapsulates the overall meaning or core of all the local characteristics. To create global features from the local features, one can run an aggregation function or operation along the rows of the 28-row local features matrix, each of which represents a local feature.

A 3D Convolutional Neural Network (C3D) can be used to extract motion characteristics from the short video clips in the Microsoft Video Description (MSVD) dataset, which can be helpful for capturing the spatiotemporal information in motion. 3D Convolutional Neural Networks (C3D) are specialised deep learning models designed to extract motion information from video data. Using C3D for motion feature extraction, spatial-temporal information are recovered from the MSVD (Microsoft Video Description). These volumetric cubes are transformed by C3D utilising 3D convolutions. The spatiotemporal patterns in the video pictures, such as motion vectors and actions, are recognised by these three-dimensional convolutional layers during training.

*D. Model Training*

This thesis on video captioning requires multiple machine-learning architectures for different purposes. Every machine learning and deep learning model has a variety of applications. Training requires encoder and decoder models with LSTM and GRU, which help train a sequence of frames. The preprocessed data was used to produce a train set, a validation set, and a test set. The feature of the training model

Data is transmitted. In encoder-decoder models, an LSTM and GRU are used as the reservoir when the LSTM functions optimally. The final output of the encoder is fed into the decoder model, which generates captions. The training set is trained in this model.

The new dataset is used to train and evaluate the proposed video captioning models. For training and validation, an 85% split ratio is utilised, and testing is conducted on 1450 and 100 video clips, respectively. The following neural network model parameters were used: 2048 unique tokens, an encoder with 28 time steps, and a latent dimension of 512. The decoder was designed to generate output sequences with a 1500 token output vocabulary and ten-time step sequences. Our training strategy involved utilising a batch size of 320 and training the model across 100 and 40 epochs, respectively. These parameter settings influenced the model's design and training regimen, which in turn impacted the model's performance within the research context.

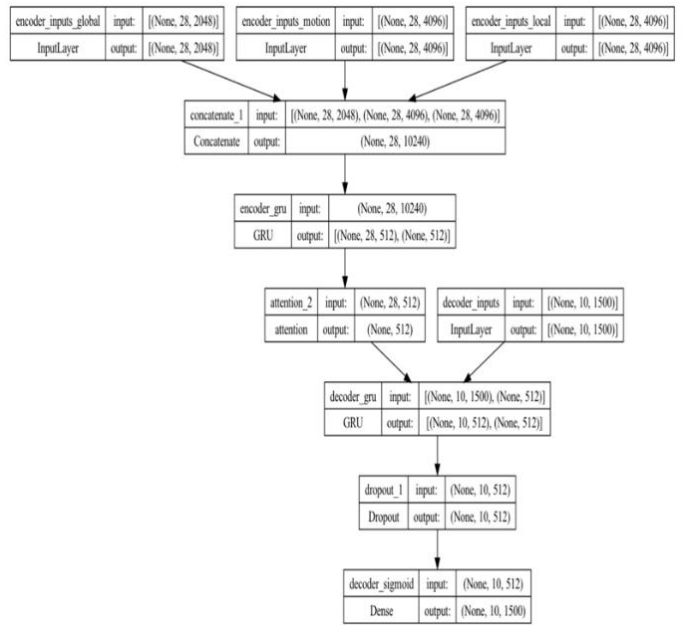

**Fig** 9: Final Model

Architecture *D. Evaluation metrics*

The four different automated evaluation metrics that are used are METEOR [22], CIDEr [23], ROUGE [24], and BLEU [22]. Several studies claim that the BLEU score may differ from the human evaluation of the generated captions due to the dynamic structure, various content pieces, events, and activities of videos. However, we used BLEU in addition to METEOR for caption evaluation after observing its recent success in MT output evaluation.

## IV. RESULT AND ANALYSIS

While extensive research has been done on captioning pictures in Nepali, not nearly as much has been done on captioning videos in the same language. Videos can be related to photos since they are made up of an ongoing sequence of images. The performance of the proposed models is also contrasted with the findings of the study on Nepali picture captioning. The chart below compares the performance of several models with the recommended ones using different evaluation metrics. With the use of attention, both Gru and LSTM were ran for 100 and 40 epochs, yielding the best accuracy of 77.83. In the picture below, different accuracy and loss plots are displayed:

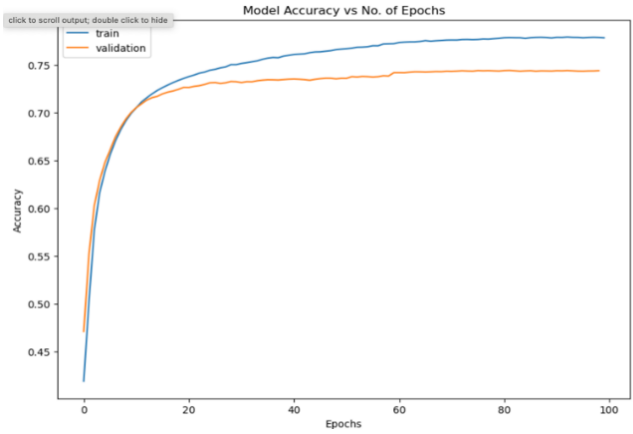

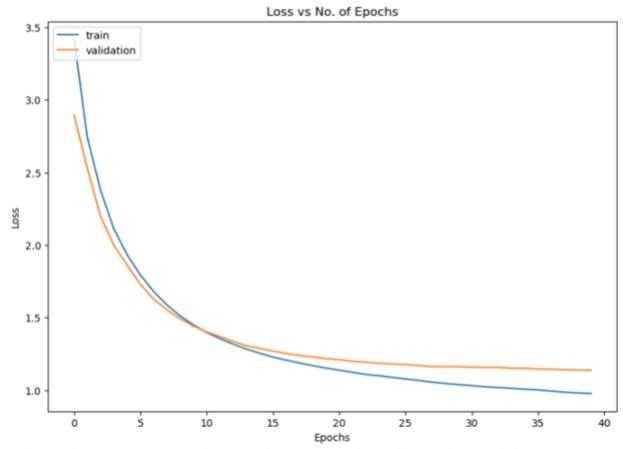

Fig 10: Accuracy loss plot encoder decoder with GRU

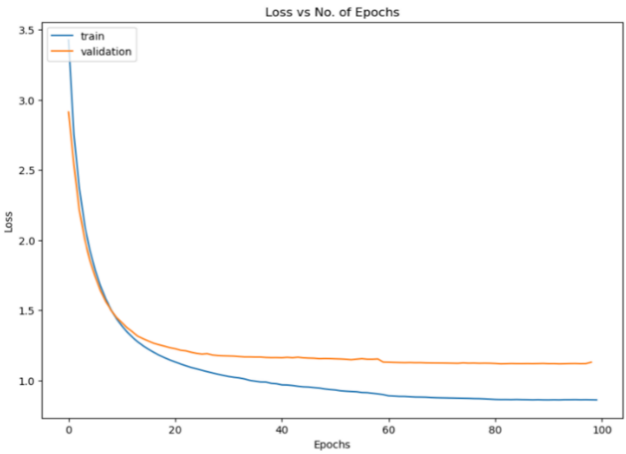

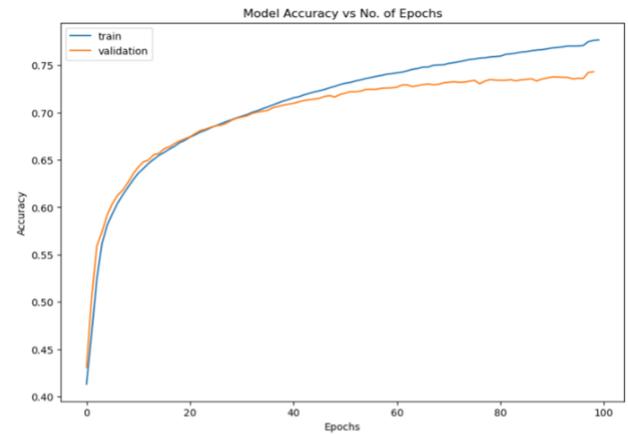

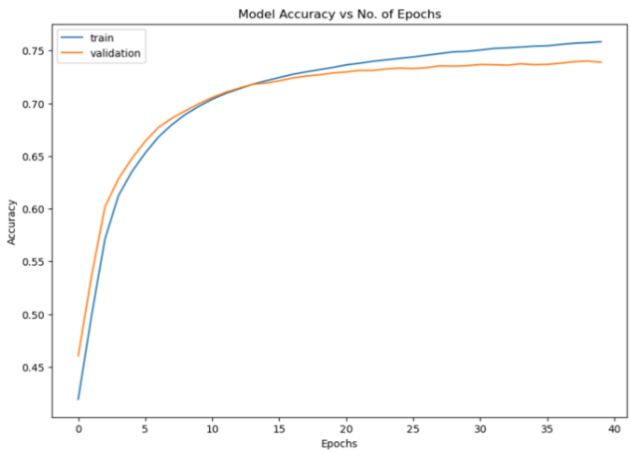

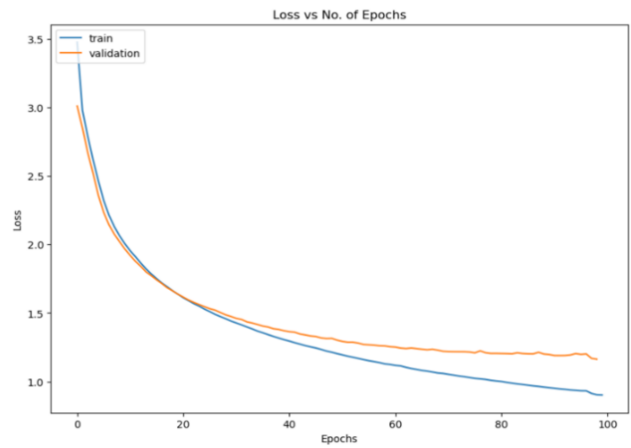

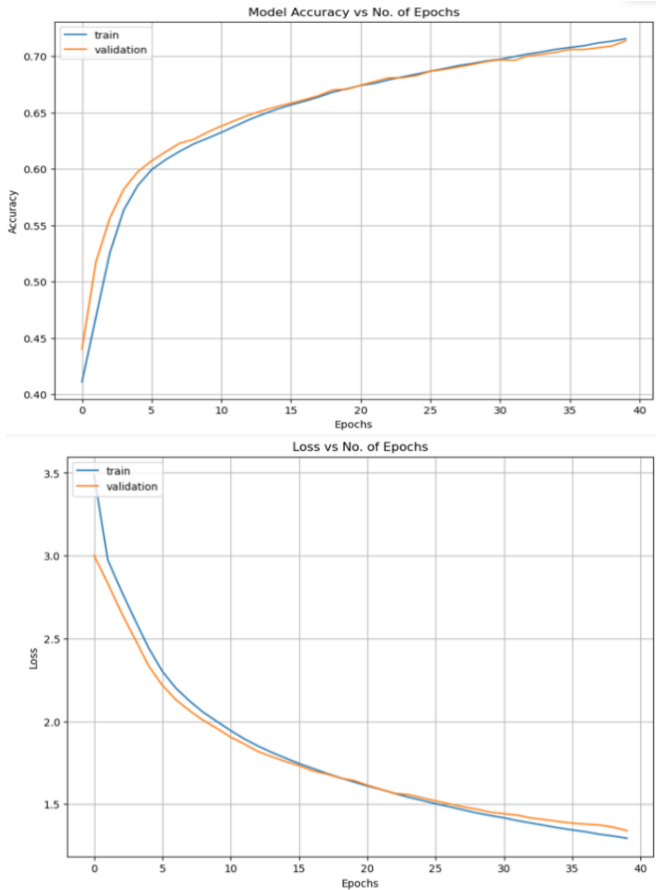

**Fig** 11: Accuracy loss plot of Encoder decoder LSTM

| RNN | Bleu1 | Bleu2 | Bleu3 | Bleu4 | METEOR | ROUGE_L |
|---|---|---|---|---|---|---|
| LSTM | 0.58 | 0.421 | 0.282 | 0.169 | 0.361 | 0.248 |
| GRU | 0.66 | 0.479 | 0.329 | 0.189 | 0.361 | 0.248 |
| LSTM+ATTENTION | 0.65 | 0.452 | 0.282 | 0.169 | 0.361 | 0.248 |
| GRU+ATTENTION | 0.68 | 0.451 | 0.282 | 0.223 | 0.361 | 0.342 |

Table 1: Performance Evaluation on Greedy Search

The exciting topic of research on Nepali video captioning has seen some recent developments. The research' findings show how crucial model selection is to achieving better performance. The GRU performs better than the LSTM in terms of performance. However, as Table shows, this research's BLEU-3 and BLEU-4 ratings were also rather low. The BLEU-4 and METEOR ratings were given particular attention in comparison to previous works. Overall, the findings imply that machine learning models' performance on caption creation tasks can be enhanced by the attention mechanism.

One reason for this could be the calibre of the dataset used for evaluation and training. The dataset was translated from English to Nepali using Google Translate; hence, there may be discrepancies and mistakes in the Nepali captions. These mistakes may cause a distortion in how well the generated captions suit the reference texts, which could lead to lower BLEU scores. The study's METEOR scores, however, were high, indicating that even with the captions' little departures from the reference lines, they are significant and successfully convey the main idea. This illustrates the model's ability to produce relevant captions in spite of the several challenges posed by the Nepali language, including word order, morphological structures, varying degrees of politeness, a lack of linguistic resources, and the potential for translation errors, to name a few. The study also found that the batch size, number of training epochs, and hidden dimensions can all have an impact on the model's performance. To look at possible enhancements, you can adjust these and other hyperparameters. Taking everything into account, the study demonstrates the feasibility and promise of Nepali video captioning. However, both the model and the descriptions might be made better.

CONCLUSION

In this work, a number of well-known neural network topologies were tested. One notable contribution of this study is the generation of a syntactically and semantically coherent dataset from the MSVD, which may be used for future research in this field. The dataset is used to train the carefully selected state-of-the-art models in order to attain the best level of accuracy in model differentiation. In addition, the attention method is introduced to obtain benchmark performance for Bengali video captioning. The optimal combination for generic RNN architecture is LSTM GRU with VGG16. Nevertheless, the attention-based GRU in conjunction with VGG16 is strong enough to generate captions that appear more realistic, making it the best model all around. The model's performance for two different search techniques is then evaluated using three commonly used evaluation metrics: BLEU, METEOR, and ROUGE, in order to undertake a thorough and flexible performance evaluation.

**ACKNOWLEDGEMENT**

We would like to thank all the professors and teachers at the Department of Electronics and Computer Engineering, Pulchowk Campus for their important and practical guidance for the research.